\documentclass[11pt]{article}
\usepackage{acl2016}
\usepackage{times}
\usepackage{latexsym}
\usepackage[centertags]{amsmath}
\usepackage{amssymb}
\usepackage{multirow}

\usepackage{graphicx}
\usepackage{varioref}
\labelformat{equation}{(#1)}

\aclfinalcopy

\DeclareMathAlphabet{\mbf}{OT1}{ptm}{b}{n}
\DeclareMathOperator*{\argsort}{arg\,sort}
\DeclareMathOperator*{\minimize}{minimize}
\def\bft{{\mbf t}}

\def\bfq{{\mbf q}}
\def\bfa{{\mbf a}}
\def\bfh{{\mbf h}}
\def\bfs{{\mbf s}}
\def\bfL{{\mbf L}}
\def\bfA{{\mbf A}}
\def\bfB{{\mbf B}}
\def\bfb{{\mbf b}}

\title{A Parallel-Hierarchical Model for Machine Comprehension on Sparse Data}

\author{Adam Trischler \\ {\tt adam.trischler} \And Zheng Ye \\ {\tt jeff.ye} \And Xingdi Yuan \\ {\tt eric.yuan} \And Jing He \\ {\tt jing.he} \And Phillip Bachman \\ {\tt phil.bachman} \AND
        Kaheer Suleman \\ {\tt k.suleman@maluuba.com} \\ Maluuba Research \\ Montreal, Qu\'{e}bec, Canada }

\date{}

\begin{document}

\maketitle

\begin{abstract}
Understanding unstructured text is a major goal within natural language processing. Comprehension tests pose questions based on short text passages to evaluate such understanding. In this work, we investigate machine comprehension on the challenging {\it MCTest} benchmark. Partly because of its limited size, prior work on {\it MCTest} has focused mainly on engineering better features. We tackle the dataset with a neural approach, harnessing simple neural networks arranged in a parallel hierarchy. The parallel hierarchy enables our model to compare the passage, question, and answer from a variety of trainable perspectives, as opposed to using a manually designed, rigid feature set. Perspectives range from the word level to sentence fragments to sequences of sentences; the networks operate only on word-embedding representations of text. When trained with a methodology designed to help cope with limited training data, our Parallel-Hierarchical model sets a new state of the art for {\it MCTest}, outperforming previous feature-engineered approaches slightly and previous neural approaches by a significant margin (over 15\% absolute).
\end{abstract}

\section{Introduction}
Humans learn in a variety of ways---by communication with each other, and by study, the reading of text. Comprehension of unstructured text by machines, at a near-human level, is a major goal for natural language processing. It has garnered significant attention from the machine learning research community in recent years.

Machine comprehension (MC) is evaluated by posing a set of questions based on a text passage (akin to the reading tests we all took in school). Such tests are objectively gradable and can be used to assess a range of abilities, from basic understanding to causal reasoning to inference~\cite{richardson2013}. Given a text passage and a question about its content, a system is tested on its ability to determine the correct answer~\cite{sachan2015}. In this work, we focus on {\it MCTest}, a complex but data-limited comprehension benchmark, whose multiple-choice questions require not only extraction but also inference and limited reasoning~\cite{richardson2013}. Inference and reasoning are important human skills that apply broadly, beyond language.

We present a parallel-hierarchical approach to machine comprehension designed to work well in a data-limited setting. There are many use-cases in which comprehension over limited data would be handy: for example, user manuals, internal documentation, legal contracts, and so on. Moreover, work towards more efficient learning from any quantity of data is important in its own right, for bringing machines more in line with the way humans learn. Typically, artificial neural networks require numerous parameters to capture complex patterns, and the more parameters, the more training data is required to tune them. Likewise, deep models learn to extract their own features, but this is a data-intensive process. Our model learns to comprehend at a high level even when data is sparse.

The key to our model is that it compares the question and answer candidates to the text using several distinct {\it perspectives}. We refer to a question combined with one of its answer candidates as a {\it hypothesis} (to be detailed below). The {\it semantic} perspective compares the hypothesis to sentences in the text viewed as single, self-contained thoughts; these are represented using a sum and transformation of word embedding vectors, similarly to in \newcite{weston2014}. The {\it word-by-word} perspective focuses on similarity matches between individual words from hypothesis and text, at various scales. As in the semantic perspective, we consider matches over complete sentences. We also use a sliding window acting on a subsentential scale (inspired by the work of~\newcite{hill2015}), which implicitly considers the linear distance between matched words. Finally, this word-level sliding window operates on two different views of text sentences: the {\it sequential} view, where words appear in their natural order, and the {\it dependency} view, where words are reordered based on a linearization of the sentence's dependency graph. Words are represented throughout by embedding vectors~\cite{mikolov2013}. These distinct perspectives naturally form a hierarchy that we depict in Figure~\ref{fig:model}. Language is hierarchical, so it makes sense that comprehension relies on hierarchical levels of understanding.

The perspectives of our model can be considered a type of feature. However, they are implemented by parametric differentiable functions. This is in contrast to most previous efforts on {\it MCTest}, whose numerous hand-engineered features cannot be trained. Our model, significantly, can be trained end-to-end with backpropagation. To facilitate learning with limited data, we also develop a unique training scheme. We initialize the model's neural networks to perform specific heuristic functions that yield decent (thought not impressive) performance on the dataset. Thus, the training scheme gives the model a safe, reasonable baseline from which to start learning. We call this technique {\it training wheels}.

Computational models that comprehend (insofar as they perform well on MC datasets) have developed contemporaneously in several research groups \cite{weston2014,sukhbaatar2015,hill2015,hermann2015,kumar2015}. Models designed specifically for {\it MCTest} include those of \newcite{richardson2013}, and more recently \newcite{sachan2015}, \newcite{wang2015mc}, and \newcite{yin2016}. In experiments, our {\it Parallel-Hierarchical} model achieves state-of-the-art accuracy on {\it MCTest}, outperforming these existing methods.

Below we describe related work, the mathematical details of our model, and our experiments, then analyze our results.

\section{The Problem}
In this section we borrow from~\newcite{sachan2015}, who laid out the MC problem nicely. Machine comprehension requires machines to answer questions based on unstructured text. This can be viewed as selecting the best answer from a set of candidates. In the multiple-choice case, candidate answers are predefined, but candidate answers may also be undefined yet restricted ({\it e.g.}, to \textsl{yes}, \textsl{no}, or any noun phrase in the text) \cite{sachan2015}.

For each question $q$, let $T$ be the unstructured text and $A = \{ a_{i} \}$ the set of candidate answers to $q$. The machine comprehension task reduces to selecting the answer that has the highest evidence given $T$. As in~\newcite{sachan2015}, we combine an answer and a question into a {\it hypothesis}, $h_i = f(q, a_i)$. To facilitate comparisons of the text with the hypotheses, we also break down the passage into sentences $t_j$, $T = \{t_j\}$. In our setting, $q$, $a_{i}$, and $t_j$ each represent a sequence of embedding vectors, one for each word and punctuation mark in the respective item.

\section{Related Work}
\label{sec:relwork}
Machine comprehension is currently a hot topic within the machine learning community. In this section we will focus on the best-performing models applied specifically to {\it MCTest}, since it is somewhat unique among MC datasets (see Section~\ref{sec:exp}). Generally, models can be divided into two categories: those that use fixed, engineered features, and neural models. The bulk of the work on {\it MCTest} falls into the former category.

Manually engineered features often require significant effort on the part of a designer, and/or various auxiliary tools to extract them, and they cannot be modified by training. On the other hand, neural models can be trained end-to-end and typically harness only a single feature: vector-representations of words. Word embeddings are fed into a complex and possibly deep neural network which processes and compares text to question and answer. Among deep models, mechanisms of attention and working memory are common, as in~\newcite{weston2014} and \newcite{hermann2015}.

\subsection{Feature-engineering models}
\newcite{sachan2015} treated {\it MCTest} as a structured prediction problem, searching for a latent {\it answer-entailing} structure connecting question, answer, and text. This structure corresponds to the best latent alignment of a hypothesis with appropriate snippets of the text. The process of (latently) selecting text snippets is related to the attention mechanisms typically used in deep networks designed for MC and machine translation~\cite{bahdanau2014,weston2014,hill2015,hermann2015}. The model uses event and entity coreference links across sentences along with a host of other features. These include specifically trained word vectors for synonymy; antonymy and class-inclusion relations from external database sources; dependencies and semantic role labels. The model is trained using a latent structural SVM extended to a multitask setting, so that questions are first classified using a pretrained top-level classifier. This enables the system to use different processing strategies for different question categories. The model also combines question and answer into a well-formed statement using the rules of~\newcite{cucerzan2005}.

Our model is simpler than that of~\newcite{sachan2015} in terms of the features it takes in, the training procedure (stochastic gradient descent {\it vs.} alternating minimization), question classification (we use none), and question-answer combination (simple concatenation or mean {\it vs.} a set of rules).

\newcite{wang2015mc} augmented the baseline feature set from~\newcite{richardson2013} with features for syntax, frame semantics, coreference chains, and word embeddings. They combined features using a linear latent-variable classifier trained to minimize a max-margin loss function. As in~\newcite{sachan2015}, questions and answers are combined using a set of manually written rules. The method of~\newcite{wang2015mc} achieved the previous state of the art, but has significant complexity in terms of the feature set.

Space does not permit a full description of all models in this category, but see also~\newcite{smith2015} and \newcite{narasimhan2015}.

Despite its relative lack of features, the Parallel-Hierarchical model improves upon the feature-engineered state of the art for {\it MCTest} by a small amount (about 1\% absolute) as detailed in Section~\ref{sec:exp}.

\subsection{Neural models}
Neural models have, to date, performed relatively poorly on {\it MCTest}. This is because the dataset is sparse and complex.

\newcite{yin2016} investigated deep-learning approaches concurrently with the present work. They measured the performance of the Attentive Reader~\cite{hermann2015} and the Neural Reasoner~\cite{peng2015}, both deep, end-to-end recurrent models with attention mechanisms, and also developed an attention-based convolutional network, the HABCNN. Their network operates on a hierarchy similar to our own, providing further evidence of the promise of hierarchical perspectives. Specifically, the HABCNN processes text at the sentence level and the \textsl{snippet} level, where the latter combines adjacent sentences (as we do through an $n$-gram input). Embedding vectors for the question and the answer candidates are combined and encoded by a convolutional network. This encoding modulates attention over sentence and snippet encodings, followed by maxpooling to determine the best matches between question, answer, and text. As in the present work, matching scores are given by cosine similarity. The HABCNN also makes use of a question classifier.

Despite the shared concepts between the HABCNN and our approach, the Parallel-Hierarchical model performs significantly better on {\it MCTest} (more than 15\% absolute) as detailed in Section~\ref{sec:exp}. Other neural models tested in~\newcite{yin2016} fare even worse.

\section{The Parallel-Hierarchical Model}
\label{sec:model}
\begin{figure}
	\centering
  	\includegraphics[width=3in]{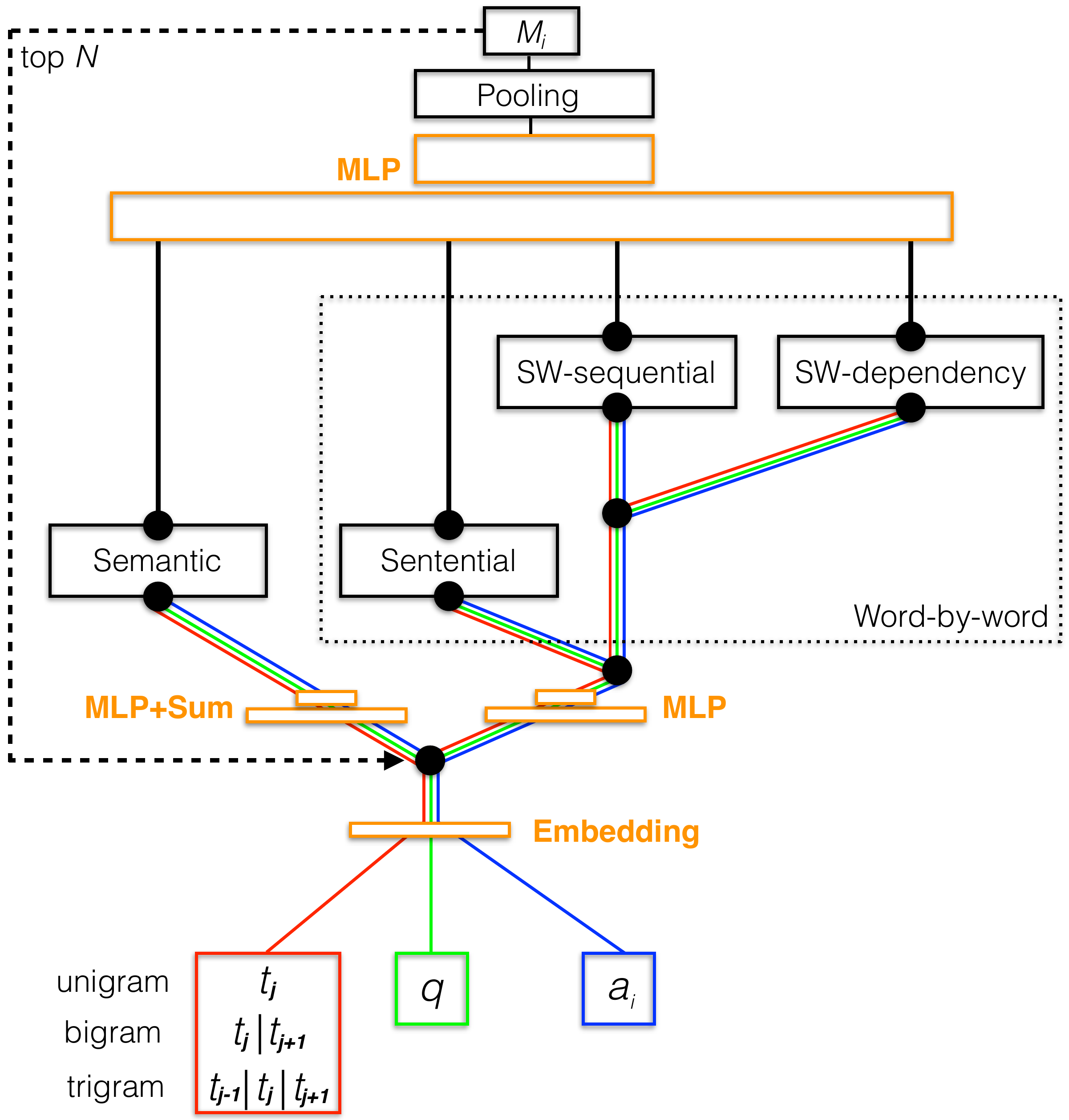}
  	\caption{ Schematic of the Parallel-Hierarchical model. SW stands for ``sliding window.'' MLP represents a general neural network.}
  	\label{fig:model}
\end{figure}

Let us now define our machine comprehension model in full. We first describe each of the perspectives separately, then describe how they are combined. Below, we use subscripts to index elements of sequences, like word vectors, and superscripts to indicate whether elements come from the text, question, or answer. In particular, we use the subscripts $k, m, n, p$ to index sequences from the text, question, answer, and hypothesis, respectively, and superscripts $t, q, a, h$. We depict the model schematically in Figure~\ref{fig:model}.

\subsection{Semantic Perspective}
The semantic perspective is similar to the Memory Networks approach for embedding inputs into memory space~\cite{weston2014}. Each sentence of the text is a sequence of $d$-dimensional word vectors: $t_j = \{\bft_k \} $, $\bft_k \in \mathbb{R}^d$. The semantic vector $\bfs^t$ is computed by embedding the word vectors into a $D$-dimensional space using a two-layer network that implements weighted sum followed by an affine tranformation and a nonlinearity; {\it i.e.},
\begin{equation}
	\bfs^t = f \left( \bfA^t \sum_k \omega_k\bft_k + \bfb_A^t \right).
	\label{eq:semantic_text}
\end{equation}
The matrix $\bfA^t \in \mathbb{R}^{D \times d}$, the bias vector $\bfb_A^t \in \mathbb{R}^{D}$, and for $f$ we use the {\it leaky ReLU} function. The scalar $\omega_k$ is a trainable weight associated to each word in the vocabulary. These scalar weights implement a kind of {\it exogenous} or bottom-up attention that depends only on the input stimulus~\cite{mayer2004}. They can, for example, learn to perform the function of stopword lists in a soft, trainable way, to nullify the contribution of unimportant filler words.

The semantic representation of a hypothesis is formed analogously, except that we combine the question word vectors $\bfq_m$ and answer word vectors $\bfa_n$ as a single sequence $\{\bfh_p\} = \{\bfq_m, \bfa_n\}$. For semantic vector $\bfs^h$ of the hypothesis, we use a unique transformation matrix $\bfA^h \in \mathbb{R}^{D \times d}$ and bias vector $\bfb_A^h \in \mathbb{R}^{D}$.

These transformations map a text sentence and a hypothesis into a common space where they can be compared. We compute the semantic match between text sentence and hypothesis using the cosine similarity,
\begin{equation}
	M^\text{sem} = \cos(\bfs^t, \bfs^h).
	\label{eq:match-semantic}
\end{equation}

\subsection{Word-by-Word Perspective}
The first step in building the word-by-word perspective is to transform word vectors from a text sentence, question, and answer through respective neural functions. For the text, $\tilde{\bft}_k = f\left( \bfB^t \bft_k + \bfb_B^t \right)$, where $\bfB^t \in \mathbb{R}^{D \times d}$, $\bfb_B^t \in \mathbb{R}^{D}$ and $f$ is again the {\it leaky ReLU}. We transform the question and the answer to $\tilde{\bfq}_m$ and $\tilde{\bfa}_n$ analogously using distinct matrices and bias vectors. In contrast with the semantic perspective, we keep the question and answer candidates separate in the word-by-word perspective. This is because matches to answer words are inherently more important than matches to question words, and we want our model to learn to use this property.

\subsubsection{Sentential}
Inspired by the work of~\newcite{wang2015snli} in paraphrase detection, we compute matches between hypotheses and text sentences at the word level. This computation uses the cosine similarity as before:
\begin{align}
	c_{km}^q &= \cos(\tilde{\bft}_k, \tilde{\bfq}_m),
	\label{eq:wbw-sentence-q}\\
	c_{kn}^a &= \cos(\tilde{\bft}_k, \tilde{\bfa}_n).
	\label{eq:wbw-sentence-a}
\end{align}

The word-by-word match between a text sentence and question is determined by taking the maximum over $k$ (finding the text word that best matches each question word) and then taking a weighted mean over $m$ (finding the average match over the full question):
\begin{equation}
	M^q = \frac{1}{Z} \sum_m \omega_m \max_k c_{km}^q.
	\label{eq:match-wbw-q}
\end{equation} 
Here, $\omega_m$ is the word weight for the question word and $Z$ normalizes these weights to sum to one over the question.
We define the match between a sentence and answer candidate, $M^a$, analogously. 
Finally, we combine the matches to question and answer according to
\begin{equation}
\label{eq:combination}
	M^\text{word} = \alpha_1 M^q + \alpha_2 M^a  + \alpha_3 M^q M^a.
\end{equation}
Here the $\alpha$ are trainable parameters that control the relative importance of the terms.

\subsubsection{Sequential Sliding Window}
The sequential sliding window is related to the original {\it MCTest} baseline by~\newcite{richardson2013}. Our sliding window decays from its focus word according to a Gaussian distribution, which we extend by assigning a trainable weight to each location in the window. This modification enables the window to use information about the distance between word matches; the original baseline used distance information through a predefined function.

The sliding window scans over the words of the text as one continuous sequence, without sentence breaks.
Each window is treated like a sentence in the previous subsection, but we include a location-based weight $\lambda(k)$. This weight is based on a word's position in the window, which, given a window, depends on its global position $k$. The cosine similarity is adapted as
\begin{align}
	s_{km}^q &= \lambda(k) \cos(\tilde{\bft}_k, \tilde{\bfq}_m),
\end{align}
for the question and analogously for the answer. We initialize the location weights with a Gaussian and fine-tune them during training.
The final matching score, denoted as $M^\text{sws}$, is computed as in~\ref{eq:match-wbw-q} and~\ref{eq:combination} with $s_{km}^q$ replacing $c_{km}^q$.

\subsubsection{Dependency Sliding Window}
The dependency sliding window operates identically to the linear sliding window, but on a different view of the text passage. The output of this component is $M^\text{swd}$ and is formed analogously to $M^\text{sws}$.

The dependency perspective uses the Stanford Dependency Parser~\cite{chen2014} as an auxiliary tool. Thus, the dependency graph can be considered a fixed feature. Moreover, linearization of the dependency graph, because it relies on an eigendecomposition, is not differentiable. However, we handle the linearization in data preprocessing so that the model sees only reordered word-vector inputs. 

Specifically, we run the Stanford Dependency Parser on each text sentence to build a dependency graph. This graph has $n_w$ vertices, one for each word in the sentence. From the dependency graph we form the Laplacian matrix $\bfL \in \mathbb{R}^{n_w \times n_w}$ and determine its eigenvectors. The second eigenvector ${\mbf u}_2$ of the Laplacian is known as the {\it Fiedler} vector. It is the solution to the minimization
\begin{equation}
	\minimize_g \sum_{i,j=1}^N \eta_{ij}(g(v_i) - g(v_j))^2,
	\label{eq:fiedler}
\end{equation}
where $v_i$ are the vertices of the graph, and $\eta_{ij}$ is the weight of the edge from vertex $i$ to vertex $j$~\cite{golub2012}. The Fiedler vector maps a weighted graph onto a line such that connected nodes stay close, modulated by the connection weights.\footnote{We experimented with assigning unique edge weights to unique relation types in the dependency graph. However, this had negligible effect. We hypothesize that this is because dependency graphs are trees, without cycles.} This enables us to reorder the words of a sentence based on their proximity in the dependency graph. The reordering of the words is given by the ordered index set
\begin{equation}
	I = \argsort({\mbf u}_2).
\end{equation}

To give an example of how this works, consider the following sentence from {\it MCTest} and its dependency-based reordering:
\begin{quote}
\textsl{Jenny, Mrs. Mustard 's helper, called the police.} \\
\textsl{the police, called Jenny helper, Mrs. 's Mustard.}
\end{quote}
Sliding-window-based matching on the original sentence will answer the question \textsl{Who called the police?} with \textsl{Mrs. Mustard}. The dependency reordering enables the window to determine the correct answer, \textsl{Jenny}. 

\subsection{Combining Distributed Evidence}
It is important in comprehension to synthesize information found throughout a document. {\it MCTest} was explicitly designed to ensure that it could not be solved by lexical techniques alone, but would instead require some form of inference or limited reasoning~\cite{richardson2013}.  It therefore includes questions where the evidence for an answer spans several sentences.

To perform synthesis, our model also takes in $n$-grams of sentences, {\it i.e.}, sentence pairs and triples strung together. The model treats these exactly as it does single sentences, applying all functions detailed above. A later pooling operation combines scores across all $n$-grams (including the single-sentence input). This is described in the next subsection.

With $n$-grams, the model can combine information distributed across contiguous sentences. In some cases, however, the required evidence is spread across distant sentences. To give our model some capacity to deal with this scenario, we take the top $N$ sentences as scored by all the preceding functions, and then repeat the scoring computations viewing these top $N$ as a single sentence.

The reasoning behind these approaches can be explained well in a probabilistic setting. If we consider our similarity scores to model the likelihood of a text sentence given a hypothesis, $p(t_j | h_i)$, then the $n$-gram and top $N$ approaches model a joint probability $p(t_{j_1},  t_{j_2}, \ldots, t_{j_k}| h_i)$. We cannot model the joint probability as a product of individual terms (score values) because distributed pieces of evidence are likely not independent.

\subsection{Combining Perspectives}
We use a multilayer perceptron to combine $M^\text{sem}$, $M^\text{word}$, $M^\text{swd}$, and $M^\text{sws}$ as a final matching score $M_i$ for each answer candidate. This network also pools and combines the separate $n$-gram scores, and uses a linear activation function.

Our overall training objective is to minimize the ranking loss
\begin{equation}
	\mathcal{L}(T, q, A) = \max(0, \mu + \max_i M_{i \neq i^\ast} - M_{i^\ast}),
	\label{eq:loss}
\end{equation}
where $\mu$ is a constant margin, $i^\ast$ indexes the correct answer, and we take the maximum over $i$ so that we are ranking the correct answer over the best-ranked incorrect answer (of which there are three). This approach worked better than comparing the correct answer to the incorrect answers individually as in~\newcite{wang2015mc}.

Our implementation of the Parallel-Hierarchical model, using the {\it Keras} framework, is available on \texttt{Github}.\footnote{http://www.hiddenwebsite.com}

\subsection{Training Wheels}
Before training, we initialized the neural-network components of our model to perform sensible heuristic functions. Training did not converge on the small {\it MCTest} without this vital approach.

Empirically, we found that we could achieve above 50\% accuracy on {\it MCTest} using a simple sum of word vectors followed by a dot product between the question sum and the hypothesis sum. Therefore, we initialized the network for the semantic perspective to perform this sum, by initializing $\bfA^x$ as the {\it identity} matrix and $\bfb_A^x$ as the zero vector, $x \in \{t,h\}$. Recall that the activation function is a $ReLU$ so that positive outputs are unchanged.

We also found basic word-matching scores to be helpful, so we initialized the word-by-word networks likewise. The network for perspective-combination was initialized to perform a sum of individual scores, using a {\it zero} bias-vector and a weight matrix of {\it ones}, since we found that each perspective contributed positively to the overall result.

This {\it training wheels} approach is related to other techniques from the literature. For instance,~\newcite{le2015} proposed the identity-matrix initialization in the context of recurrent neural networks in order to preserve the error signal through backpropagation. In residual networks~\cite{he2015}, shortcut connections bypass certain layers in the network so that a simpler function can be trained in conjunction with the full model.

\section{Experiments}
\label{sec:exp}
\subsection{The Dataset}
{\it MCTest} is a collection of 660 elementary-level children's stories and associated questions, written by human subjects. The stories are fictional, ensuring that the answer must be found in the text itself, and carefully limited to what a young child can understand~\cite{richardson2013}.

The more challenging variant consists of 500 stories with four multiple-choice questions each. Despite the elementary level, stories and questions are more natural and more complex than those found in synthetic MC datasets like {\it bAbI}~\cite{weston2014} and {\it CNN}~\cite{hermann2015}.



{\it MCTest} is challenging because it is both complicated and small. As per~\newcite{hill2015}, ``it is very difficult to train statistical models only on {\it MCTest}.'' Its size limits the number of parameters that can be trained, and prevents learning any complex language modeling simultaneously with the capacity to answer questions.

\subsection{Training and Model Details}
In this section we describe important details of the training procedure and model setup. For a complete list of hyperparameter settings, our stopword list, and other minutiae, we refer interested readers to our \texttt{Github} repository.

For word vectors we use Google's publicly available embeddings, trained with \texttt{word2vec} on the 100-billion-word {\it News corpus}~\cite{mikolov2013}. These vectors are kept fixed throughout training, since we found that training them was not helpful (likely because of {\it MCTest}'s size). The vectors are 300-dimensional ($d = 300$).

We do not use a stopword list for the text passage, instead relying on the trainable word weights to ascribe global importance ratings to words. These weights are initialized with the inverse document frequency (IDF) statistic computed over the {\it MCTest} corpus.\footnote{We override the IDF initialization for words like \textsl{not}, which are frequent but highly informative.} However, we do use a short stopword list for questions. This list nullifies query words such as \{\textsl{Who, what, when, where, how}\}, along with conjugations of the verbs \textsl{to do} and \textsl{to be}.

Following earlier methods, we use a heuristic to improve performance on negation questions~\cite{sachan2015,wang2015mc}. 
When a question contains the words \textsl{which} and \textsl{not}, we negate the hypothesis ranking scores so that the minimum becomes the maximum.

The most important technique for training the model was the {\it training wheels} approach. Without this, training was not effective at all. The identity initialization requires that the network weight matrices are square ($d=D$).

We found {\it dropout}~\cite{srivastava2014} to be particularly effective at improving generalization from the training to the test set, and used $0.5$ as the dropout probability. Dropout occurs after all neural-network transformations, if those transformations are allowed to change with training. Our best performing model held networks at the word-by-word level fixed.

For combining distributed evidence, we used up to trigrams over sentences and our best-performing model reiterated over the top two sentences ($N=2$).

We used the {\it Adam} optimizer with the standard settings~\cite{kingma2014} and a learning rate of 0.003. To determine the best hyperparameters we performed a grid search over 150 settings based on validation-set accuracy. {\it MCTest}'s original validation set is too small for reliable hyperparameter tuning, so, following~\newcite{wang2015mc}, we merged the training and validation sets of {\it MCTest}-160 and {\it MCTest}-500, then split them randomly into a 250-story training set and a 200-story validation set.

\subsection{Results}
Table~\ref{tab:results} presents the performance of feature-engineered and neural methods on the {\it MCTest} test set. Accuracy scores are divided among questions whose evidence lies in a single sentence ({\it single}) and across multiple sentences ({\it multi}), and among the two variants. Clearly, {\it MCTest}-160 is easier.

The first three rows represent feature-engineered methods. \newcite{richardson2013} + RTE is the best-performing variant of the original baseline published along with {\it MCTest}. It uses a lexical sliding window and distance-based measure, augmented with rules for recognizing textual entailment. We described the methods of~\newcite{sachan2015} and~\newcite{wang2015mc} in Section~\ref{sec:relwork}. On {\it MCTest}-500, the Parallel Hierarchical model significantly outperforms these methods on {\it single} questions ($>2\%$) and slightly outperforms the latter two on {\it multi} questions ($\approx 0.3\%$) and overall ($\approx 1\%$). The method of ~\newcite{wang2015mc} achieves the best overall result on {\it MCTest}-160. We suspect this is because our neural method suffered from the relative lack of training data.

The last four rows in Table~\ref{tab:results} are neural methods that we discussed in Section~\ref{sec:relwork}. Performance measures are taken from~\newcite{yin2016}. Here we see our model outperforming the alternatives by a large margin across the board ($>15\%$). The Neural Reasoner and the Attentive Reader are large, deep models with hundreds of thousands of parameters, so it is unsurprising that they performed poorly on {\it MCTest}. The specifically-designed HABCNN fared better, its convolutional architecture cutting down on the parameter count. Because there are similarities between our model and the HABCNN, we hypothesize that much of the performance difference is attributable to our {\it training wheels} methodology.

\begin{table*}[t]
	\small
	\centering
    \begin{tabular}{ | c || c | c | c || c | c | c |}
    	\hline
    	\multirow{2}{*}{Method} & \multicolumn{3}{|c|}{{\it MCTest}-160 accuracy (\%)} & \multicolumn{3}{|c|}{{\it MCTest}-500 accuracy (\%)} \\ \cline{2-7}
    	 & Single (112) & Multiple (128) & All & Single (272) & Multiple (328) & All \\ \hline \hline
    	 Richardson et al. (2013) + RTE & 76.78 & 62.50 & 69.16 & 68.01 & 59.45 & 63.33 \\
    	 Sachan et al. (2015) & - & - & - & 67.65 & 67.99 & 67.83 \\
    	 Wang et al. (2015) & \textbf{84.22} & 67.85 & \textbf{75.27} & 72.05 & 67.94 & 69.94 \\ \hline
    	 Attentive Reader & 48.1 & 44.7 & 46.3 & 44.4 & 39.5 & 41.9 \\
    	 Neural Reasoner & 48.4 & 46.8 & 47.6 & 45.7 & 45.6 & 45.6 \\
    	 HABCNN-TE & 63.3 & 62.9 & 63.1 & 54.2 & 51.7 & 52.9 \\
		 Parallel-Hierarchical & 79.46 & \textbf{70.31} & 74.58 & \textbf{74.26} & \textbf{68.29} & \textbf{71.00} \\
		\hline
    \end{tabular}
    \caption{Experimental results on {\it MCTest}.}
	\label{tab:results}
\end{table*}

\section{Analysis and Discussion}
We measure the contribution of each component of the model by ablating it. Results are given in Table~\ref{tab:ablate}.
\begin{table}
	\small
	\centering
    \begin{tabular}{ | c | c |}
    	\hline
    	Ablated component & Test accuracy (\%) \\ \hline \hline
    	- & \textbf{71.00} \\ \hline
    	$n$-gram & 66.51 \\ \hline
    	Top $N$ & 70.34 \\ \hline
    	Sentential & 64.33 \\ \hline
    	SW-sequential & 68.00 \\ \hline
    	SW-dependency & 70.00 \\ \hline
    	Word weights & 66.51 \\ \hline
    \end{tabular}
    \caption{Ablation study on {\it MCTest}-500 (all).}
	\label{tab:ablate}
\end{table}
Not surprisingly, the $n$-gram functionality is important, contributing almost 5\% accuracy improvement. Without this, the model has almost no means for synthesizing distributed evidence.
The top $N$ function contributes very little to the overall performance, suggesting that most {\it multi} questions have their evidence distributed across contiguous sentences.
Ablating the sentential component made the most significant difference, reducing performance by more than 5\%. Simple word-by-word matching is obviously useful on {\it MCTest}.
The sequential sliding window makes a 3\% contribution, highlighting the importance of word-distance measures.
On the other hand, the dependency-based sliding window makes only a minor contribution. We found this surprising. It may be that linearization of the dependency graph removes too much of its information.
Finally, the exogenous word weights make a significant contribution of almost 5\%.

Analysis reveals that most of our system's test failures occur on questions about quantity ({\it e.g.}, \textsl{How many...?}) and temporal order ({\it e.g.}, \textsl{Who was invited last?}). Quantity questions make up 9.5\% of our errors on the validation set, while order questions make up 10.3\%. This weakness is not unexpected, since our architecture lacks any capacity for counting or tracking temporal order. Incorporating mechanisms for these forms of reasoning is a priority for future work (in contrast, the Memory Network model is quite good at temporal reasoning~\cite{weston2014}).

The Parallel-Hierarchical model is simple. It does no complex language or sequence modeling. Its simplicity is a response to the limited data of {\it MCTest}. Nevertheless, the model achieves state-of-the-art results on the {\it multi} questions, which (putatively) require some limited reasoning. Our model is able to handle them reasonably well just by stringing important sentences together. Thus, the model imitates reasoning with a heuristic. This suggests that, to learn true reasoning abilities, {\it MCTest} is too simple a dataset---and it is almost certainly too small for this goal.

However, it may be that human language processing can be factored into separate processes of {\it comprehension} and {\it reasoning}. If so, the Parallel-Hierarchical model is a good start on the former. Indeed, if we train the method exclusively on {\it single} questions then its results become even more impressive: we can achieve a test accuracy of \textbf{79.1\%} on {\it MCTest}-500.

\section{Conclusion}
We have presented the novel Parallel-Hierarchical model for machine comprehension, and evaluated it on the small but complex {\it MCTest}. Our model achieves state-of-the-art results, outperforming several feature-engineered and neural approaches.

Working with our model has emphasized to us the following (not necessarily novel) concepts, which we record here to promote further empirical validation.
\begin{itemize}
	\item Good comprehension of language is supported by hierarchical levels of understanding ({\it Cf.}~\newcite{hill2015}).
	\item Exogenous attention (the trainable word weights) may be broadly helpful for NLP.
	\item The {\it training wheels} approach, that is, initializing neural networks to perform sensible heuristics, appears helpful for small datasets.
	\item {\it Reasoning} over language is challenging, but easily simulated in some cases.
\end{itemize}

\bibliography{mcrefs}
\bibliographystyle{acl2016}

\end{document}